\documentclass[a4paper]{elsarticle}
\usepackage{graphicx}
\usepackage{graphicx}
\usepackage{color}
\usepackage{lineno,hyperref}
\usepackage{import}
\usepackage{algorithmic}
\usepackage{algorithm}
\usepackage{amsmath,amssymb,amsfonts}
\usepackage{movie15}
\usepackage{multirow}
\usepackage{booktabs}

\begin{document}

\begin{frontmatter}

\title{An Interactive Vision–Language Platform for Cognitive Remediation in Schizophrenia}

\author{Nassira AIT MEHDI, Milissa TEMMAM, Slimane LARABI}

\address{RIIMA Laboratory, Computer Science Faculty, \\
USTHB University, 16111, Algeria}

\begin{abstract}

Cognitive remediation tasks often require patients to perform structured actions involving object manipulation and sequential reasoning. For patients diagnosed with schizophrenia, these tasks are crucial for addressing severe cognitive deficits, including executive dysfunction, impaired working memory, and diminished visuospatial reasoning. However, evaluating the correctness of these physical actions generally relies on manual observation by clinicians, which introduces subjectivity and limits the scalability of therapeutic interventions. In this paper, we propose an automated framework based on Vision–Language Models (VLMs) for action verification in cognitive remediation tasks tailored for schizophrenia rehabilitation.

The proposed system relies on a camera-monitored tabletop environment composed of structured miniature scenes including roads, a roundabout, a park, and toy vehicles. Patients receive audio instructions describing goal-oriented spatial actions to perform by manipulating a toy vehicle. These interactive physical activities are specifically designed to stimulate targeted cognitive functions, such as sustained attention, motor coordination, spatial navigation, and cognitive flexibility, in an ecologically valid setting.

To verify the correctness of the performed actions without requiring continuous clinical oversight, the system analyzes the video feed tracking the patient's hand and toy movements. A fine-tuned Vision–Language Model interprets the recorded video sequences and generates semantic descriptions of the observed activities, enabling high-level verification of the executed actions with respect to the initial textual instructions. A dedicated dataset of 4,634 tabletop cognitive remediation video scenarios was collected to evaluate the proposed approach. Experimental results demonstrate that our specialized framework effectively bridges low-level physical telemetry with high-level clinical feedback, presenting a scalable and objective solution for advanced cognitive rehabilitation.

\end{abstract}

\begin{keyword}
Vision–Language Models (VLM) \sep Cognitive Training Environments \sep Human–Computer Interaction \sep Action Recognition.   \end{keyword}

\end{frontmatter}

\section{Introduction}

Mental health disorders represent one of the most pressing challenges in global public health today. Among these, schizophrenia stands out as a severe condition characterized not only by clinical symptoms but also by profound, persistent cognitive deficits. Patients with schizophrenia frequently experience significant executive dysfunction, impaired visuospatial working memory, and diminished action-planning capabilities. These deficits create a substantial barrier to daily functioning, preventing patients from organizing sequential tasks, navigating environments effectively, and maintaining independent living.

Cognitive Remediation Techniques (CRT) refer to a set of structured therapeutic tools designed to target these specific impairments. This strictly non-pharmacological approach is used in addition to standard psychiatric care and does not replace it. Its goal is to restore impaired cognitive functions or compensate for persistent deficits by stimulating neuroplasticity and overcoming cognitive obstacles [5]. Traditionally, CRT relies heavily on abstract paper-and-pencil tasks or repetitive digital computer games. While these methods can improve isolated cognitive metrics, they often suffer from a major clinical limitation: low ecological validity. Patients frequently struggle to transfer the abstract skills learned on a screen or paper into the complex, physical, and multi-sensory interactions required in real-world scenarios.

To bridge this gap, we propose an interactive tabletop platform that provides a physical, ecologically valid setting tailored for cognitive rehabilitation in schizophrenia. The system utilizes a structured miniature environment composed of roads, a roundabout, and a parking area, where patients physically manipulate a toy vehicle to accomplish complex, multi-step goal-directed tasks. This tangible setup is uniquely suited for treating the specific deficits of schizophrenia. By requiring patients to execute sequential navigation constraints in a physical 3D space, it forces the re-engagement of real-world spatial planning, motor coordination, and executive control mechanisms. Automating the evaluation of these physical interactions, however, requires a system capable of bridging low-level motion data with high-level semantic reasoning to verify task execution accurately without constant clinical oversight.

This paper begins with a review of related work on existing intelligent systems for the remediation of brain disorders. In Section 3, we present the proposed approach, including dataset collection, as well as the training and testing of the employed VLM. Section 4 presents the results obtained.

\section{Related Works}

Cognitive Remediation Therapy (CRT) aims to improve key cognitive functions in individuals with cognitive impairments. 
It can be broadly divided into two main approaches: paper-and-pencil tasks and computer-assisted training.

Paper-and-pencil tasks include structured exercises targeting core cognitive functions such as attention, working memory, problem-solving, executive functions, visuospatial abilities, and cognitive flexibility. These tasks aim to improve focus, reasoning, planning, spatial understanding, and adaptability to changing conditions, thereby supporting daily cognitive functioning.

Computer-assisted training has been widely explored using artificial intelligence, virtual reality, and human–computer interaction techniques. While these systems improve accessibility, they may suffer from limited engagement and lack of standardization ~\cite{Quan2024}. irtual reality systems provide immersive, personalized environments that enhance cognitive functions like memory and attention~\cite{Quan2024, Fiorillo2026}; however, they remain limited by small-scale studies and methodological variability. AI-driven and gamified systems enable adaptive feedback and improved engagement ~\cite{El-Banna2026}, but often lack generalization and unified modeling. Adaptive and AR-based approaches adjust task difficulty based on performance ~\cite{ManChu2025}, yet typically rely on simplified evaluation strategies.\\ 
Brain–computer interfaces ~\cite{Zheng2023} and robotic systems enable real-time interaction and monitoring; however, most existing approaches still rely on predefined metrics and lack high-level semantic understanding for automated action evaluation. Crucially, few of these digital frameworks address the specific need of schizophrenia patients to practice physical, spatially-anchored sequence planning that translates directly to daily living activities.

Despite the advancements in both manual and digital Cognitive Remediation Therapy (CRT) frameworks, several fundamental limitations persist across the literature. First, CRT lacks a universally standardized definition, which introduces significant heterogeneity in therapeutic protocols and clinical objectives across different studies. This variance is further compounded by discrepancies in intervention durations and assessment methodologies, heavily compromising the reproducibility of clinical outcomes. Second, existing frameworks lack a reliable, automated mechanism to verify whether a patient has accurately executed a prescribed physical task. Consequently, most contemporary systems still necessitate continuous, direct clinician supervision and offer minimal automated interpretation of patient behavior, limiting their scalability and objectivity in clinical settings.
 
To address the limitations of existing systems, we propose an interactive platform for cognitive remediation based on vision–language models which provide significant added value compared to current cognitive rehabilitation approaches by overcoming several fundamental shortcomings of traditional solutions which rely either on digital exercises or virtual reality environments, without a direct link to the patient’s behavior in a real physical setting. 

\section{The Proposed Interactive Platform}

\subsection{Material platform: The Description}

\begin{figure}
\includegraphics[height=5cm]{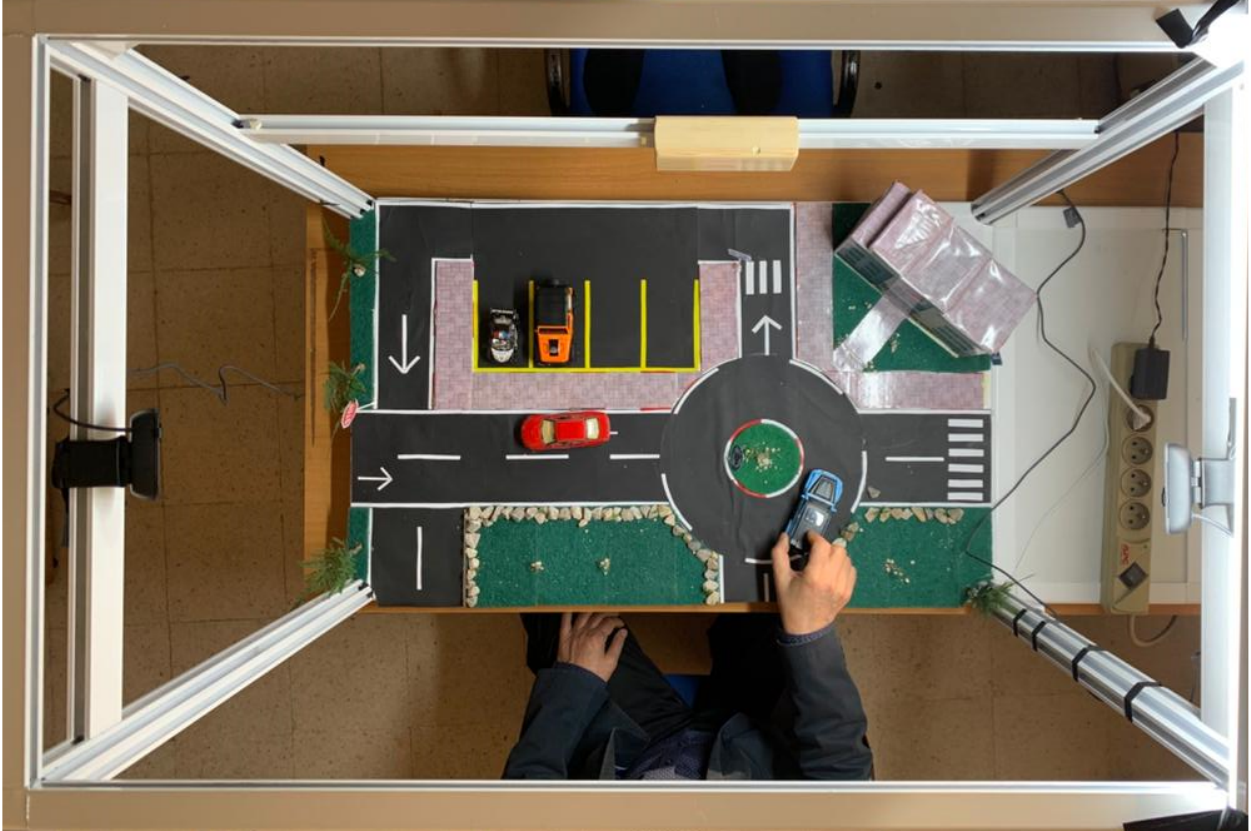}
\includegraphics[height=5cm]{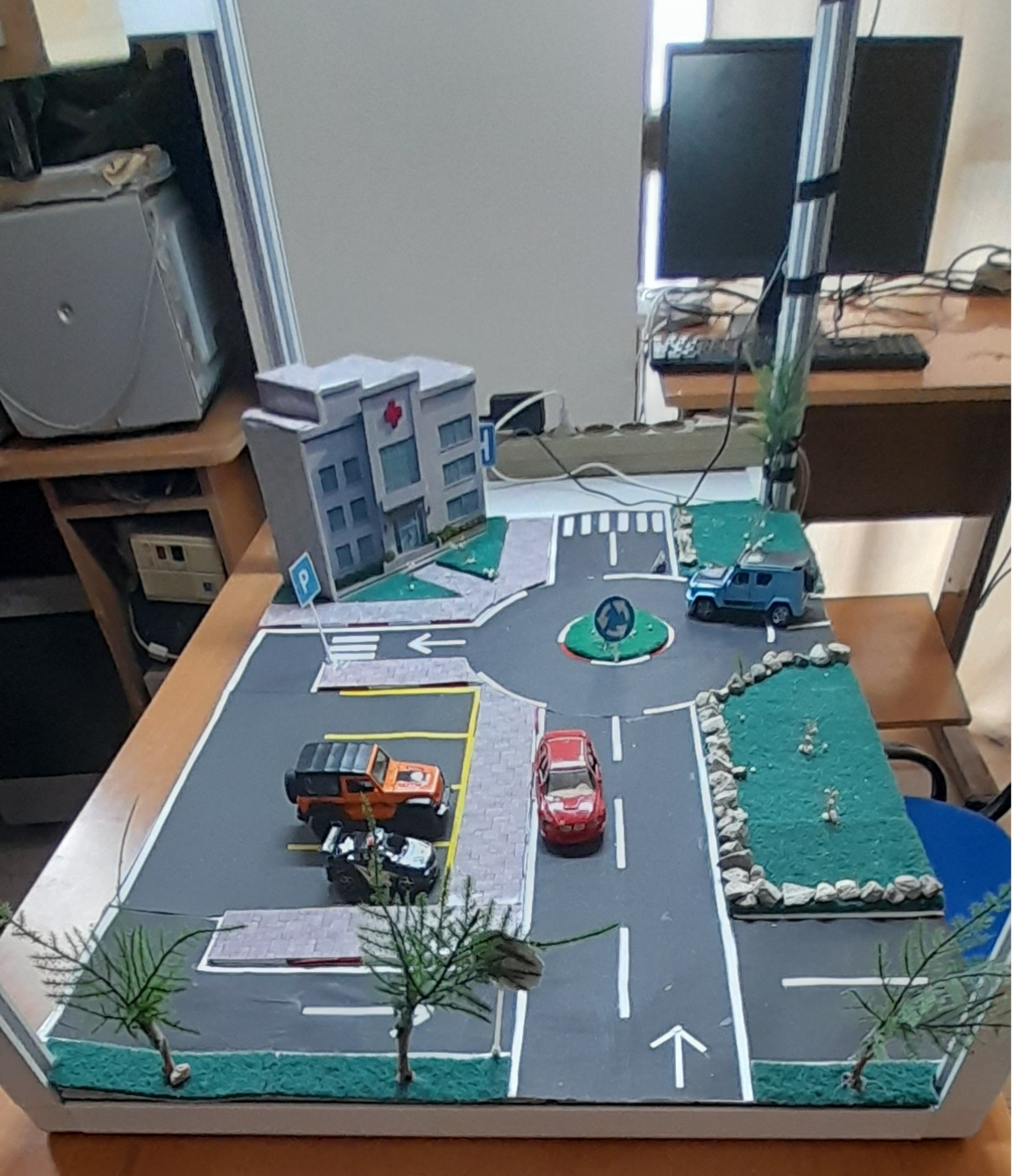}
\caption{The realized platform (left: top view ), (right: side view).} \label{fig1}
\end{figure}

Our interactive platform is composed of two parts: The first component is a physical setup consisting of a table monitored by two cameras, as illustrated in Fig.~\ref{fig1}. The table contains a three-dimensional scene corresponding to specific therapeutic protocol defined by the clinician, in which patients are asked to perform a set of activities.
The illustrated scene is composed of roads, a roundabout, a parking area, and toy vehicles. The second component is an intelligent analysis module that processes video streams of user interactions. It uses Vision–Language Models for semantic interpretation of actions. The system evaluates whether the observed behavior is consistent with both spatial constraints and textual instructions, enabling automatic action verification and feedback generation.

At the beginning of each exercise, the system provides the patient with verbal instructions through audio output. The instruction describes a simple scenario involving physical interaction with objects placed on a table, sometimes including specific constraints. The patient has to listen, understand, and then act accordingly. As example of such instructions: "Take the car, move it to the park, and park it, but go via the roundabout". 
The patient performs the task in a real environment, interacting with physical objects placed on a table. This setup makes the exercise more realistic, the patient is not limited to screen-based abstract tasks, but manipulating real objects in a real space.

\subsection{Understanding action using Vision Language Model}

The action of the patient is recorded and passed to our AI model which generates textual description of what the patient is performing.
This description is then automatically compared to the ground truth to determine whether the patient performed it properly or not.
The exercises are defined by the clinician through a dedicated PC interface, allowing a personalized progression adapted to each patient's level and condition. The clinician can adjust the complexity of the scenario, the number of objects, the constraints, and the expected sequence of actions.
The solution can be applied to any type of remediation exercise defined by the clinician, whether concentration exercises, memory tasks, problem-solving activities, or executive function training. Any scenario involving physical actions on a table can be integrated into the system.
At the end of each session, a complete report is automatically produced and made available to the clinician, providing an objective and detailed assessment of the patient's performance and allowing them to track progress over time.

\subsubsection{Vision Language Model Selection}

Recent advances in Vision Language Models (VLMs) have significantly transformed video understanding and action recognition in unconstrained environments. Earlier approaches primarily relied on task-specific architectures combining 2D or 3D convolutional neural networks with recurrent networks or temporal attention mechanisms. Although these methods achieved strong performance in action classification, they remain limited in their ability to generate rich semantic descriptions of observed actions.

More recently, multimodal foundation models such as GPT-4V ~\cite{openai2023}, Gemini ~\cite{google2023}, and instruction-tuned video-language models, including Video-LLaMA ~\cite{Zhang2023} and VideoChatGPT ~\cite{maaz2023}, have enabled open-ended video reasoning and natural language description generation. These models exhibit strong generalization capabilities across diverse visual domains; however, they often face challenges in fine-grained temporal reasoning and domain-specific action recognition.

In parallel, structured reasoning approaches, including 3DVLA ~\cite{zhen2024}, GraphCoTVLA 
 ~\cite{huang2026}, and GPT-4Scene ~\cite{qi2025}, have introduced explicit spatial reasoning and graph-based representations for 3D scene understanding and action modeling. While these methods improve spatio-temporal reasoning, they generally require complex processing pipelines and additional perception modules.

In this work, we adopt \textbf{Qwen2.5-VL-3B-Instruct} ~\cite{Bai2025} as the backbone of our video captioning framework. Qwen2.5-VL is a multimodal foundation model that jointly processes visual and textual information, enabling effective understanding of sequential video frames and their associated temporal dynamics. The model provides a favorable balance between multimodal reasoning capability, computational efficiency, and adaptability to parameter-efficient fine-tuning. Furthermore, its native support for video inputs and instruction-following makes it well suited for generating detailed descriptions of cognitive rehabilitation exercises.

The \textbf{3B-Instruct} variant was selected instead of the larger 7B model for two main reasons:

\begin{itemize}
    \item \textbf{Computational Efficiency:} The 3B model substantially reduces GPU memory requirements and training time, enabling efficient fine-tuning and deployment on standard cloud infrastructures while maintaining competitive performance.

    \item \textbf{Dataset Scale:} The model capacity is well matched to the size and complexity of the proposed dataset (4,634 annotated videos covering 17 rehabilitation exercises). A smaller model reduces the risk of overfitting while providing sufficient capacity to learn domain-specific visual and semantic representations.
\end{itemize}


\subsubsection{Fine-Tuning Framework: LLaMA-Factory}

The downstream adaptation of the vision-language model was performed using LLaMA-Factory, an open-source framework for efficient large language model fine-tuning. LLaMA-Factory was selected because of its native support for the Qwen2.5-VL architecture and its seamless integration with multimodal datasets represented in JSON format. In addition, the framework provides built-in support for parameter-efficient fine-tuning (PEFT), advanced attention mechanisms, thereby simplifying the implementation of large-scale multimodal fine-tuning while reducing memory consumption ~\cite{Zheng2024}.


\subparagraph{Memory Optimization with LoRA}

To enable efficient training on commodity GPU hardware and prevent out-of-memory (OOM) errors, Low-Rank Adaptation (LoRA) was adopted as the parameter-efficient fine-tuning strategy. Rather than updating all parameters of the Qwen2.5-VL model, LoRA freezes the pretrained weights and introduces trainable low-rank matrices into the linear layers. Consequently, only a small fraction of the model parameters is optimized during training, substantially reducing GPU memory requirements while preserving the pretrained visual and linguistic representations.

\subsubsection{Fine-Tuning}

\textbf{Stage 1: Action Anchoring}
The first training stage consisted of supervised fine-tuning (SFT) using the structured action annotations. The objective of this stage was to establish robust visual-semantic associations between video sequences and their corresponding action labels. Learning these structured representations enables the model to acquire domain-specific visual knowledge before being exposed to more complex natural language descriptions during the subsequent training stage.

\textbf{Stage 2: Natural Language Refinement}
Stage 2 builds upon the action-anchored model checkpoint produced by Stage 1. The objective transitions from structured token prediction to generating fluid, contextualized natural language descriptions of clinical exercises. The model is fine-tuned using the natural language annotations.

\subsubsection{Semantic Comparison Module}

Once the video captioning model generates a textual description of the patient's actions, the system must determine whether the generated description is semantically consistent with the clinician-defined ground truth.

The ground truth consists of a natural language description written by the clinician during exercise creation. It specifies the expected sequence of actions when the exercise is performed correctly. This reference description is stored in the database together with the exercise and serves as the basis for evaluating the generated caption.

Since the generated caption and the reference description may use different wording while conveying the same meaning, an exact lexical comparison is insufficient. Instead, a semantic comparison is required to assess the similarity between the two descriptions. This problem is commonly addressed using \emph{Semantic Textual Similarity} (STS), which measures the degree of semantic equivalence between two texts by assigning a numerical similarity score~\cite{agirre2016}. Existing STS methods range from traditional lexical approaches and supervised machine learning techniques to modern deep learning models based on sentence embeddings. In embedding-based approaches, each sentence is represented as a dense vector in a semantic space, and their similarity is typically computed using cosine similarity. The resulting score ranges from $-1$ to $1$, where $1$ denotes identical semantic meaning, $0$ indicates no semantic relationship, and $-1$ represents opposite meanings, although negative values are rarely observed in practice.

To compute the semantic similarity between the generated caption $C_g$ and the clinician's reference description $R_c$, we employ Sentence-BERT (SBERT)~\cite{Reimers2019}. SBERT maps each textual string to a dense, fixed-dimensional vector embedding $\mathbf{v} \in \mathbb{R}^d$. The semantic equivalence between the two text segments is subsequently quantified using the cosine similarity of their respective embedding vectors:

\begin{equation}
\text{Sim}(C_g, R_c) = \frac{\mathbf{v}_{C_g} \cdot \mathbf{v}_{R_c}}{\|\mathbf{v}_{C_g}\| \|\mathbf{v}_{R_c}\|}
\end{equation}

The resulting score yields a continuous metric where values approaching $1$ represent strict semantic alignment, allowing the framework to successfully match paraphrastic variations that conventional token-matching algorithms would erroneously penalize.
Although the similarity score provides an overall measure of semantic agreement, it does not explain the specific differences between the two descriptions. Therefore, a large language model is incorporated to generate an interpretable evaluation of the comparison results. In our framework, \textbf{LLaMA~3} is used to analyze the generated caption, the reference description, and the SBERT similarity score, producing a detailed explanation that highlights correctly recognized actions, omitted information, and semantic discrepancies. Owing to its strong multilingual reasoning and text generation capabilities, LLaMA~3 is well suited for this semantic evaluation task.

To preserve data confidentiality and avoid transmitting patient information to external cloud services, the semantic evaluation module is deployed locally using Ollama, which provides an efficient framework for running large language models on local hardware by managing model weights, dependencies, and inference through a unified interface. Consequently, all semantic analyses are performed locally, ensuring data privacy while eliminating the latency and operational costs associated with cloud-based APIs.

\section{Experimental Study}
\subsection{Dataset Collection and Clinical Formulation}

Videos were recorded within a simulated 3D environment designed for cognitive remediation protocols. While the experimental sandbox evaluated in this study features a virtual driving environment, it acts as a representative surrogate for complex procedural testing. In a deployment scenario, a clinician configures the specific parameters of the 3D scene and dictates the behavioral objectives to be performed by patients diagnosed with schizophrenia, targeting deficits in executive functioning, spatial orientation, and action planning. 

To ensure the diversity and robustness of the evaluation framework, 4,634 action videos were captured across varying simulation states. These included structural variations in hand appearance, vehicle type, targeted action trajectories, and camera perspectives. The average duration of the recorded tasks is 4 to 5 seconds, processed in MP4 format under landscape orientation. 

Each video sequence is mapped to two distinct annotations:
\begin{enumerate}
    \item \textbf{Simple Annotation (Descriptive):} A short, highly structured domain description optimized for automated validation scripts and direct classification tasks.
    \item \textbf{Natural Annotation (Natural Language Description):} A fluid, contextualized narrative utilized during Stage 2 fine-tuning to transform rigid behavioral telemetry into feedback that is intelligible by the VLM model.
\end{enumerate}

The overall task repository is organized into five functional operational categories, each grouping specific traffic actions designed to challenge cognitive control mechanisms~\cite{larabi2026cognitiveremediation}. Table~\ref{tab1} provides a comprehensive summary of the dataset's structural configuration.

 Table~\ref{tab1} gives a summary of all categories and actions.
\begin{table}
\caption{Dataset structure — categories, actions, size and number of videos.}
\label{tab1}
\begin{tabular}{|l|l|p{4cm}|l|}
\hline
Category & Size (GB) & Action & Videos \\
\hline
\multirow{4}{*}{Parking} 
& \multirow{4}{*}{21.2 GB} 
& Park the car correctly within the designated parking lines. & 770 \\
\cline{3-4}
& 
& Park the car incorrectly by placing it randomly outside the designated parking spaces. & 61 \\
\cline{3-4}
& 
& Park the car incorrectly so that it is not aligned with the parking lines and occupies two parking spaces. & 313 \\
\cline{3-4}
& 
& Park on the sidewalk. & 263 \\
\hline
\multirow{4}{*}{Roundabout} 
& \multirow{4}{*}{23.7 GB} 
& Turn around the roundabout correctly (correct direction, full loop). & 108 \\
\cline{3-4}
& 
& Turn around the roundabout in the wrong direction. & 95 \\
\cline{3-4}
& 
& Completes only part of the roundabout. & 529 \\
\cline{3-4}
& 
& Cross directly the roundabout. & 199 \\
\hline
\multirow{4}{*}{\begin{tabular}{c}
Parking \\
Entrance, Exit
\end{tabular}} 
& \multirow{4}{*}{18.1 GB} 
& Enter through the entrance. & 174 \\
\cline{3-4}
& 
& Exit through the exit. & 180 \\
\cline{3-4}
& 
& Enter through the exit. & 177  \\
\cline{3-4}
& 
& Exit through the entrance. & 175 \\
\hline
Sidewalk & 15.6 GB & Driving on the sidewalk.	& 653 \\
\hline
\multirow{2}{*}{Roads} 
& \multirow{2}{*}{20.3 GB} 
& Drive in correct lane. & 483 \\
\cline{3-4}
& 
& Drive in wrong direction. & 454 \\
\hline
\end{tabular}
\end{table}

\subsection{Video Captioning Model Evaluation}
\label{sec:evaluation}

\subsubsection{Evaluation Methodology}
\label{subsec:eval_methodology}

The quantitative assessment of all model variants encompassing Stage 1, Stage 2, and the pretrained baseline was conducted using a unified, multi-metric evaluation pipeline. This pipeline computes seven standard natural language generation metrics designed to capture complementary dimensions of caption quality: BLEU-1, BLEU-2, BLEU-3, BLEU-4, ROUGE-L, METEOR, and CIDEr.

To ensure strict comparability across all three model variants, the evaluation adhered to the following protocol:\\

\begin{itemize}
    \item Validation Split: The held-out validation set was reconstructed identically for all configurations using the Hugging Face \textit{Dataset.train\_test\_split()} utility with \textit{test\_size=0.1}, \textit{seed=42}, and \textit{shuffle=True}. This step precisely mirrors the internal data partitioning utilized by LLaMA-Factory during training, eliminating potential data leakage and ensuring all models are evaluated on the exact same 463 video samples.

    \item Evaluation Script: A unified evaluation script was executed for Stage 1, Stage 2, and the baseline model. This guarantees absolute consistency across generation parameters (e.g., temperature, top-p), tokenization preprocessing pipelines, and downstream metric computations.
\end{itemize}
During both training stages, the model was optimized using standard \textbf{Cross-Entropy Loss} tailored for causal language modeling . This objective guides the Qwen2.5-VL model through a next-token prediction task, minimizing the negative log-likelihood of the target token sequence ~\cite{Bai2025}.

\subsubsection{Stage 1 Evaluation Results}
\label{subsec:stage1_results}

The evolution of training and validation losses during Stage 1 fine-tuning provides critical insight into the model's ability to anchor visual inputs to the structured action taxonomy while avoiding overfitting.

\begin{figure}[H]
    \centering
    \includegraphics[width=0.90\textwidth]{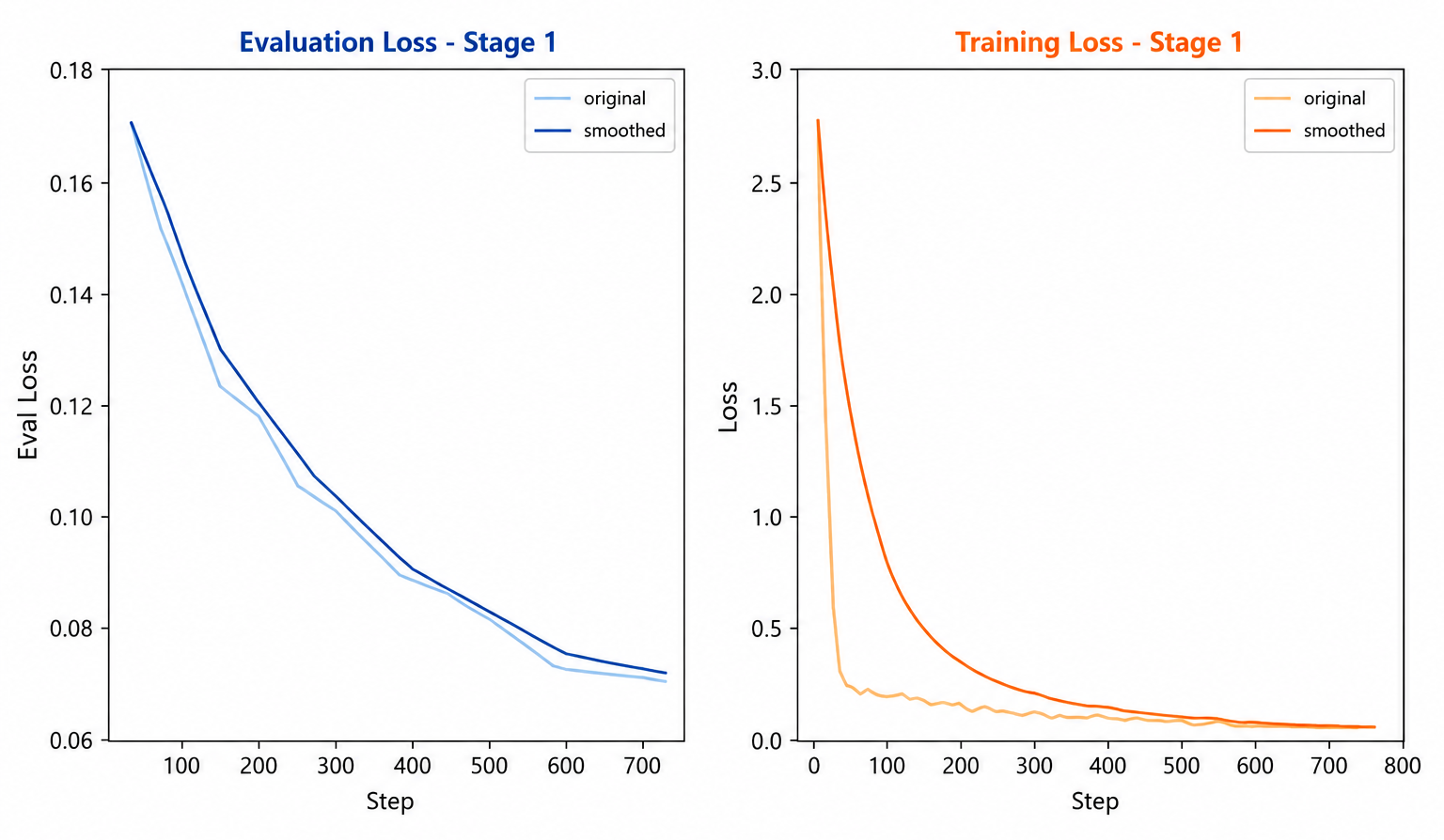}
    \caption{Stage 1 training and validation loss curves.}
    \label{fig:stage1_loss}
\end{figure}

Figure~\ref{fig:stage1_loss} presents the training and validation loss curves for Stage 1, illustrating the convergence behavior.

The Stage 1 training and evaluation loss curves demonstrate healthy parallel convergence. The training loss drops sharply from an initial value of approximately 2.7 to a final plateau near 0.02. The evaluation loss follows a similar trajectory, decreasing from $\approx$0.165 to $\approx$0.067. Both curves decrease monotonically and stabilize in tandem, with the validation loss remaining well above the training loss throughout.

This pattern provides decisive evidence against overfitting. In a typical overfitting scenario, the training loss would continue to decrease while the validation loss plateaus or increases --- a divergence entirely absent here. The persistent gap between training and evaluation loss ($\approx$0.05 at convergence) is expected and healthy, reflecting the natural difficulty gap between seen training samples and unseen validation samples.

No underfitting is detected either. The rapid descent to near-zero training loss confirms that the LoRA rank of 32, combined with the 3-epoch budget and learning rate of $1\times10^{-4}$, provides sufficient capacity to capture the structured, template-based annotation distribution of Stage 1.

The Stage 1 model, fine-tuned on structured, token-based action annotations, was evaluated on the 463-sample validation set. The results are summarized in Table~\ref{tab:stage1_results}.

\begin{table}[H]
\centering
\caption{Stage 1 Evaluation Metrics (Action Anchoring)}
\label{tab:stage1_results}
\begin{tabular}{lr}
\toprule
\textbf{Metric} & \textbf{Score} \\
\midrule
BLEU-1   & 88.36 \\
BLEU-2   & 85.46 \\
BLEU-3   & 82.87 \\
BLEU-4   & 80.48 \\
METEOR   & 88.62 \\
ROUGE-1  & 89.67 \\
ROUGE-2  & 83.32 \\
ROUGE-L  & 89.20 \\
CIDEr    & 699.21 \\
\bottomrule
\end{tabular}
\end{table}

The Stage 1 results demonstrate exceptional performance across all metric categories. The BLEU scores exhibit a remarkably narrow decay from BLEU-1 (88.36) to BLEU-4 (80.48), with a gap of only 7.88 points. This indicates that the model does not merely match isolated keywords but successfully reproduces long, coherent phrase structures present in the reference annotations.

The METEOR score of 88.62 confirms strong semantic alignment, as METEOR's synonym-aware matching captures paraphrastic variations that strict n-gram metrics would penalize. ROUGE-1 (89.67) and ROUGE-L (89.20) are nearly identical, demonstrating that the longest common subsequence between predictions and references spans almost the entire annotation.

The CIDEr score of 699.21 is exceptionally high, reflecting the structural homogeneity of the Stage 1 annotations where near-identical templates are assigned to videos depicting the same clinical action. CIDEr's TF-IDF weighting mechanism heavily rewards this intra-class repetition, amplifying the score beyond typical general-domain ranges (MS COCO: $\approx$1.0--1.5) ~\cite{vedantam2015cider}. While this confirms the model's ability to consistently reproduce clinically specific terms, the elevation is primarily a dataset property rather than an indicator of superior linguistic diversity.

\subsubsection{Stage 2 Evaluation Results}
\label{subsec:stage2_results}

The evolution of training and validation losses during Stage 2 fine-tuning reveals how the model progressively refines its fluency on natural language descriptions while maintaining convergence stability.

\begin{figure}[H]
    \centering
    \includegraphics[width=0.90\textwidth]{"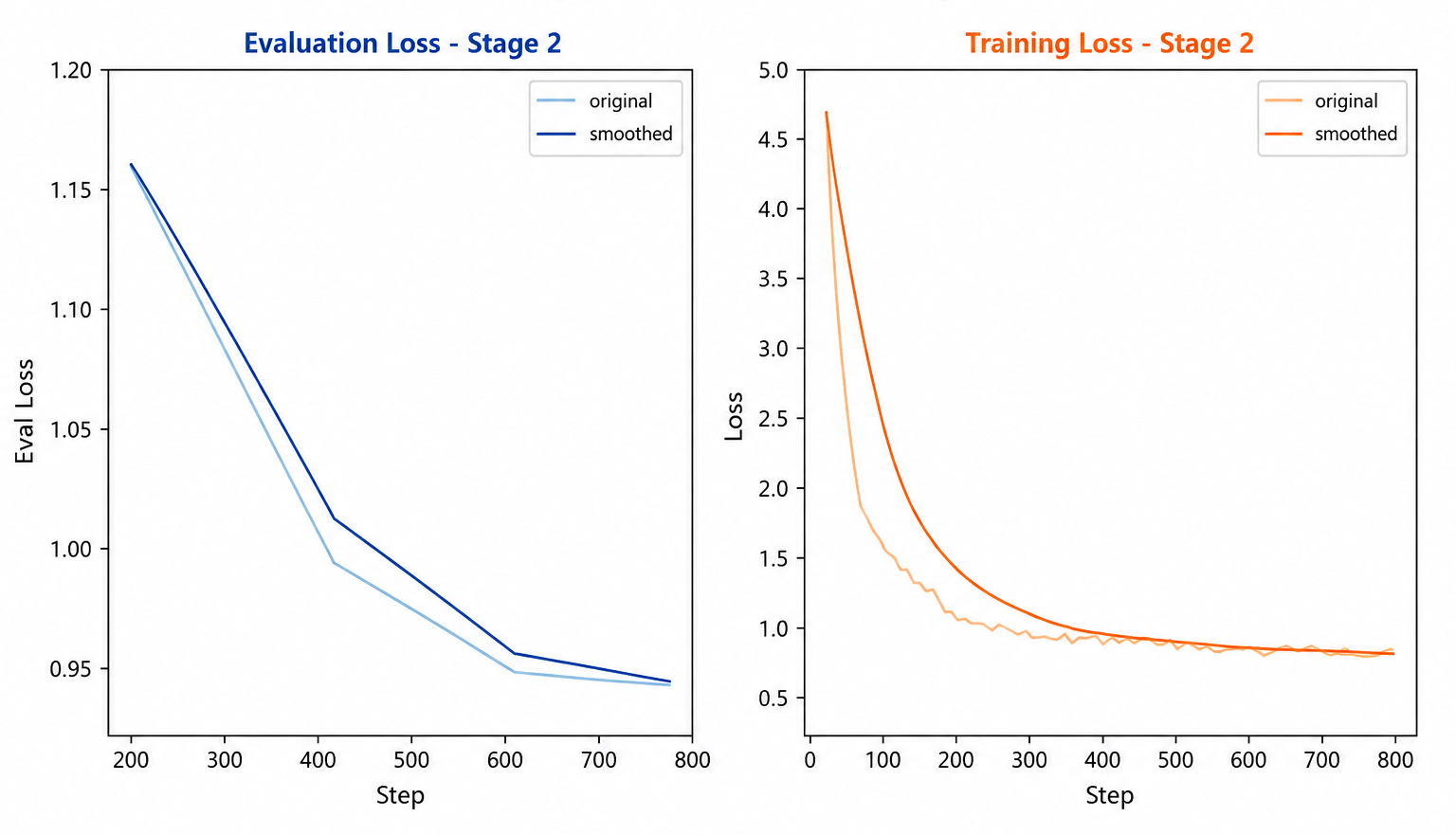"}
    \caption{Stage 2 training and validation loss curves.}
    \label{fig:stage2_loss}
\end{figure}

Figure~\ref{fig:stage2_loss} presents the training and validation loss curves for Stage 2, illustrating the convergence behavior.

The Stage 2 training and evaluation loss curves exhibit qualitatively similar convergence behavior to Stage 1, but with notably higher absolute values reflecting the increased task complexity. The training loss begins at approximately 4.6 --- substantially higher than Stage 1's $\approx$2.7 --- and decreases monotonically to a final value near 0.95. The evaluation loss starts at $\approx$1.16 and converges to $\approx$0.94.

The tight final alignment between training loss ($\approx$0.95) and evaluation loss ($\approx$0.94) is particularly significant. Both curves decrease in parallel without any upward divergence of the validation loss, providing strong evidence against overfitting. The near-identical terminal values indicate that the model generalizes effectively from training to unseen validation data, despite the much higher initial losses compared to Stage 1.

The higher absolute loss values are expected and diagnostically meaningful. Natural language generation is inherently more complex than template-based classification: the model must learn a broader distribution of syntactic structures, vocabulary choices, and contextual variations. That the training loss stabilizes at $\approx$0.95 rather than near-zero confirms the model is learning a genuinely harder task.

No underfitting is present. The monotonic decrease and stable convergence demonstrate that the Stage 2 configuration (LoRA rank 32, learning rate $1\times10^{-5}$, 3 epochs, DDP across dual L40 GPUs) provides adequate representational capacity for the natural language refinement task. The reduced learning rate successfully prevents catastrophic forgetting of Stage 1's anchored visual-semantic associations while allowing sufficient plasticity for linguistic adaptation.

The Stage 2 model was evaluated on the identical 463-sample validation set using the same protocol. The results are summarized in Table ~\ref{tab:stage2_results}.

\begin{table}[H]
\centering
\caption{Stage 2 Evaluation Metrics (Natural Language Refinement)}
\label{tab:stage2_results}
\begin{tabular}{lr}
\toprule
\textbf{Metric} & \textbf{Score} \\
\midrule
BLEU-1   & 56.79 \\
BLEU-2   & 40.82 \\
BLEU-3   & 29.32 \\
BLEU-4   & 21.54 \\
METEOR   & 51.51 \\
ROUGE-1  & 56.28 \\
ROUGE-2  & 30.17 \\
ROUGE-L  & 50.20 \\
CIDEr    & 119.61 \\
\bottomrule
\end{tabular}
\end{table}

The Stage 2 results demonstrate an expected quantitative shift relative to Stage 1 across all metrics; however, this variation must not be misinterpreted as a reduction in model performance. The cumulative BLEU scores exhibit a natural decay from BLEU-1 (56.79\%) to BLEU-4 (21.54\%). This 35.25-point gap directly reflects the inherent structural difficulty of matching exact higher-order n-grams when the targeted ground-truth annotations transition from rigid tokens to complex natural language varying in syntax and word order.

In this context, the achieved METEOR score of 51.51\% is particularly revealing: it notably exceeds the METEOR benchmark of 42.3\% reported for the massive open-domain Qwen2-VL-72B base model on the standard MSR-VTT dataset ~\cite{xu2016msrvtt}, despite our architecture utilizing only 3B parameters. Because METEOR incorporates synonym-aware matching and stemming alignment, this high performance confirms that the model successfully generates semantically equivalent and clinically valid feedback, even when it introduces paraphrastic rewordings that rigid n-gram metrics penalize.

Similarly, the ROUGE-1 (56.28\%) and ROUGE-L (50.20\%) metrics remain remarkably close. This tight alignment demonstrates that the core chronological sequence of the clinical exercise , subject, action, spatial context, and performance outcome, is robustly preserved across the temporal window, even when surface lexical choices shift.

Finally, the achieved CIDEr score of 119.61 falls squarely within the optimal expected range for fluid natural language video captioning. Unlike the artificially inflated CIDEr values observed during Stage 1, which were skewed by the heavy repetition of short keyword templates, this normalized score reflects genuine semantic consensus with a highly diverse, descriptive reference distribution.

\section{Conclusion and Future Work}

In this paper, we proposed an interactive cognitive remediation platform that integrates a physical 3D tabletop environment with advanced semantic reasoning to objectively evaluate task execution in patients diagnosed with schizophrenia. Unlike conventional cognitive remediation systems that rely on manual observation or rigid, screen-based digital exercises, our approach leverages a specialized multi-stage fine-tuning pipeline to bridge physical clinical interactions with high-level semantic feedback. 

Experimental results demonstrate that our specialized framework successfully captures fine-grained action dynamics within realistic rehabilitation scenarios. The fine-tuned Qwen2.5-VL-3B-Instruct model effectively learned domain-specific clinical vocabulary, generating precise, action-centered descriptions aligned with therapeutic protocols. Furthermore, the local integration of an SBERT and LLaMA~3 semantic comparison module provides a privacy-preserving, objective mechanism to validate action conformity against clinician-defined ground truths. The results confirm that combining multi-frame video understanding with semantic textual similarity offers a scalable solution for automated clinical assessment without requiring continuous oversight.

Several avenues exist to extend this work. First, future research will investigate the integration of multi-modal cues—such as patient speech, gaze tracking, and facial expressions—to construct a richer, multi-dimensional assessment of a patient's executive functioning during therapy sessions. Second, we plan to validate the clinical efficacy of this platform through longitudinal studies involving target patient cohorts in active psychiatric rehabilitation settings.

\end{document}